# How Far Should We Look Back to Achieve Effective Real-Time Time-Series Anomaly Detection?

Ming-Chang Lee[1], Jia-Chun Lin[2], and Ernst Gunnar Gran[3]

[1,2,3]Department of Information Security and Communication Technology, Norwegian University of Science and Technology,
Ametyst-bygget, 2815 Gjøvik, Norway
[3]Simula Research Laboratory, 1364 Fornebu, Norway

[1] ming-chang.lee@ntnu.no
[2]jia-chun.lin@ntnu.no
[3]ernst.g.gran@ntnu.no

10th January 2023

# How Far Should We Look Back to Achieve Effective Real-Time Time-Series Anomaly Detection?

Ming-Chang Lee[1], Jia-Chun Lin[2], and Ernst Gunnar Gran[3,4]

[1,2,3] Department of Information Security and Communication Technology, Norwegian University of Science and Technology, 2815 Gjøvik, Norway
[4] Simula Research Laboratory, 1364 Fornebu, Norway
`[1]ming-chang.lee@ntnu.no`
`[2]jia-chun.lin@ntnu.no`
`[3]ernst.g.gran@ntnu.no`

**Abstract.** Anomaly detection is the process of identifying unexpected events or abnormalities in data, and it has been applied in many different areas such as system monitoring, fraud detection, healthcare, intrusion detection, etc. Providing real-time, lightweight, and proactive anomaly detection for time series with neither human intervention nor domain knowledge could be highly valuable since it reduces human effort and enables appropriate countermeasures to be undertaken before a disastrous event occurs. To our knowledge, RePAD (Real-time Proactive Anomaly Detection algorithm) is a generic approach with all abovementioned features. To achieve real-time and lightweight detection, RePAD utilizes Long Short-Term Memory (LSTM) to detect whether or not each upcoming data point is anomalous based on short-term historical data points. However, it is unclear that how different amounts of historical data points affect the performance of RePAD. Therefore, in this paper, we investigate the impact of different amounts of historical data on RePAD by introducing a set of performance metrics that cover novel detection accuracy measures, time efficiency, readiness, and resource consumption, etc. Empirical experiments based on real-world time series datasets are conducted to evaluate RePAD in different scenarios, and the experimental results are presented and discussed.

## 1  Introduction

Anomaly detection refers to any method or technique designed to identify unexpected events or anomalies in data, and it has been used to discover anomalies in a time series, which is a series of data points generated continuously and evenly indexed in time order. During the past decade, a number of approaches and methods have been introduced for time series anomaly detections such as [2-8]. Many of them require either domain knowledge (e.g., data patterns, data distributions, pre-labeled training data) or human intervention (e.g., collecting sufficient training data, pre-labelling training data, configuring appropriate hyperparameters, determining proper detection thresholds, etc.). These requirements consequently limit the applicability of these approaches.

It could be highly valuable to provide an anomaly detection approach that requires neither human intervention nor domain knowledge because such an approach can significantly reduce human effort and that it can be easily and immediately applied to detect anomalies in any time series. In addition, offering real-time and proactive anomaly detection might be also highly appreciated since it enables appropriate actions and countermeasures to be undertaken as early as possible. Last but not least, it would be beneficial to offer a lightweight anomaly detection approach because such an approach can be deployed on any commodity machine, such as desktops, laptops, and mobile phones.

To our knowledge, RePAD [1] is an approach with all above-mentioned features. RePAD is a real-time proactive anomaly detection algorithm for streaming time series based on a special type of recurrent neural networks called Long Short-Term Memory (LSTM). As soon as RePAD is employed to detect anomalies in a time series, it trains a LSTM model with short-term historical data points and automatically calculates a detection threshold to determine if each upcoming data point is anomalous or not. RePAD also decides if it should retrain its LSTM model based on the most recent short-term data points so as to adapt to minor pattern change in the time series. By utilizing LSTM with a simple network structure and training the LSTM model with short-term historical data points, RePAD demonstrates good, cost-effective, and real-time anomaly detection performance according to [1]. However, it is unclear from [1] that how different amounts of historical data points impact the performance of RePAD and that how far RePAD should look back such that it is able to provide the most effective anomaly detection. By the most effective, we mean that:

- Anomaly detect can soon start without lengthy preprocessing
- Anomalies can be detected as early as possible
- Low false positives and low false negatives
- Infrequent LSTM model retraining

In this paper, we conduct a comprehensive study to investigate the performance of RePAD. The study starts by introducing a set of metrics which cover novel detection accuracy measures that are appropriate and suitable for proactive detection approaches. The metrics also cover readiness and time efficiency since achieving lightweight and real-time anomaly detection as mentioned earlier are essential and desirable. We design two empirical experiments based on real-world time series datasets provided by the Numenta Anomaly Benchmark (NAB) [9] to evaluate RePAD in different scenarios. The experiment results show that RePAD based on long-term historical data points leads to a long preparation period, frequent LSTM retraining, and long detection time. Most importantly, RePAD loses its capability to accurately detect anomalies. On the contrary, choosing short-term historical data points is highly recommended for RePAD since it enables RePAD to provide the most effective anomaly detection.

The rest of the paper is organized as follows: Section 2 introduces the design of RePAD. Section 3 presents all metrics used to evaluate the performance of RePAD. Section 4 presents and discusses evaluation results based on empirical experiments and real-world time series datasets. In Section 5, we conclude this paper and outline future work.

## 2  RePAD

RePAD uses LSTM with a simple network structure (e.g., one hidden layer with only ten hidden units) to predict the values of the data points arriving at the next time point based on the data values observed at the past $b$ continuous time points. Note that $b$ is called the Look-Back parameter, which is set to 3 in [1]. Fig. 1 illustrates the algorithm of RePAD. Let $t$ be the current time point and $t$ starts from 0, which is the time point when RePAD is launched. Since $b$ is set to 3, RePAD has to collect three data points observed at time points 0, 1, and 2, denoted by $v_0$, $v_1$, and $v_2$, respectively. With these three data points, RePAD at time point 2 is able to train a LSTM model (called $M$) and uses $M$ to predict the next data value, denoted by $\widehat{v_3}$. After that, RePAD continues to train another LSTM model with the most three recent data points to predict $\widehat{v_4}$ and $\widehat{v_5}$, respectively at time point 3 and 4, respectively.

When $t$ advances to 5, implying that the real data-point values $v_5$ have been observed respectively, RePAD is able to calculate the corresponding AARE (i.e., $AARE_5$) based on Equation 1:

$$AARE_t = \frac{1}{b} \cdot \sum_{y=t-b+1}^{t} \frac{|v_y - \widehat{v_y}|}{v_y}, t \geq 2b-1 \tag{1}$$

where $v_y$ is the observed data value at time point $y$, and $\widehat{v_y}$ is the forecast data value for time point $y$. It is clear that $AARE_5 = \frac{1}{3} \cdot \left( \frac{|v_3 - \widehat{v_3}|}{v_3} + \frac{|v_4 - \widehat{v_4}|}{v_4} + \frac{|v_5 - \widehat{v_5}|}{v_5} \right)$. Recall that AARE stands for Average Absolute Relative Error. A low AARE value indicates that the predicted values are close to the observed values. After deriving $AARE_5$, RePAD replaces $M$ with another LSTM model trained with the most three recent data points to predict $\widehat{v_5}$. The same process repeats when $t$ advances to 6.

When $t = 7$ (i.e., $t = 2b + 1$), RePAD calculates $AARE_7$. At this moment, RePAD is able to officially detect anomalies since it has already $b$ AARE values (i.e., $AARE_5$, $AARE_6$, and $AARE_7$), which are sufficient to calculate detection threshold $thd$. In other words, RePAD requires a preparation period of $2b + 1$ time points to get its detection started. Equation 2 shows how $thd$ is calculated based on the Three-Sigma Rule [2], which is commonly used for anomaly detection.

$$thd = \mu_{AARE} + 3 \cdot \sigma, t \geq 2b+1 \tag{2}$$

where $\mu_{AARE} = \frac{1}{t-b-1} \cdot \sum_{x=2b-1}^{t} AARE_x$, where (i.e., the average AARE of all previously derived AARE values), and $\sigma$ is the corresponding standard deviation, i.e., $\sigma = \sqrt{\frac{\sum_{x=2b-1}^{t}(AARE_x - \mu_{AARE})^2}{t-b-1}}$. If $AARE_7$ is lower than or equal to $thd$ (see line 15 of Fig. 1), RePAD does not considers $v_7$ as anomalous. Hence, it keeps using $M$ to predict the next data point $\widehat{v_8}$. However, if $AARE_7$ is higher than $thd$ (i.e., line 16 holds), it means that either the data pattern of the time series has changed or an anomaly happens.

To be sure, RePAD retrains an LSTM with the most $b$ recent data points (i.e., $v_4, v_5, v_6$) to re-predict $\widehat{v_7}$ and re-calculate $AARE_7$.

**RePAD algorithm**
**Input**: Data points in a time series
**Output**: Anomaly notifications
**Procedure:**

| | |
|---|---|
| 1: | Let $t$ be the current time point and $t$ starts from 0; Let *flag* be True; |
| 2: | **While** time has advanced { |
| 3: | Collect data point $v_t$; |
| 4: | **if** $t \geq b - 1$ and $t < 2b - 1$ { // i.e., $2 \leq t < 5$, if $b = 3$ |
| 5: | Train an LSTM model by taking $[v_{t-b+1}, v_{t-b+2} \dots, v_t]$ as the training data; |
| 6: | Let $M$ be the resulting LSTM model and use $M$ to predict $\widehat{v_{t+1}}$;} |
| 7: | **else if** $t \geq 2b - 1$ and $t < 2b + 1$ { //i.e., $5 \leq t < 7$, if $b = 3$ |
| 8: | Calculate $AARE_t$ based on Equation 1; |
| 9: | Train an LSTM model by taking $[v_{t-b+1}, v_{t-b+2} \dots, v_t]$ as the training data; |
| 10: | Let $M$ be the resulting LSTM model and use $M$ to predict $\widehat{v_{t+1}}$;} |
| 11: | **else if** $t \geq 2b + 1$ and *flag*==True { //i.e., $t \geq 7$ if $b = 3$ |
| 12: | **if** $t \neq 7$ { Use $M$ to predict $\widehat{v_t}$;} |
| 13: | Calculate $AARE_t$ based on Equation 1; |
| 14: | Calculate *thd* based on Equation 2; |
| 15: | **if** $AARE_t \leq thd$ { $v_t$ is <u>not</u> considered as an anomaly;} |
| 16: | **else**{ |
| 17: | Train an LSTM model with $[v_{t-b}, v_{t-b+1}, \dots, v_{t-1}]$; |
| 18: | Use the newly trained LSTM model to predict $\widehat{v_t}$; |
| 19: | Calculate $AARE_t$ using Equation 1; |
| 20: | Calculate *thd* based on Equation 2; |
| 21: | **if** $AARE_t \leq thd$ { $v_t$ is <u>not</u> considered as an anomaly;} |
| 22: | **else** { |
| 23: | $v_t$ is reported as an anomaly immediately; |
| 24: | Let *flag* be False;}}} |
| 25: | **else if** $t \geq 2b + 1$ and *flag*==False { |
| 26: | Train an LSTM model with $[v_{t-b}, v_{t-b+1}, \dots, v_{t-1}]$; |
| 27: | Use the newly trained LSTM model to predict $\widehat{v_t}$; |
| 28: | Calculate $AARE_t$ based on Equation 1; |
| 29: | Calculate *thd* based on Equation 2; |
| 30: | **if** $AARE_t \leq thd${ |
| 31: | $v_t$ is <u>not</u> considered as an anomaly; |
| 32: | Replace $M$ with the new LSTM model from line 26; |
| 33: | Let *flag* be True;} |
| 34: | **else** { |
| 35: | $v_t$ is reported as an anomaly immediately; Let *flag* be False;}}} |

**Fig. 1.** The algorithm of RePAD.

If the new $AARE_7$ is lower than or equal to *thd* (see line 21), RePAD confirms that the data pattern of the time series has changed and that $v_7$ is not anomalous. On the contrary, if the new $AARE_7$ is still higher than *thd* (see line 22), RePAD considers $v_7$ anomalous since the LSTM trained with the most recent data points is still unable to accurately predict $v_7$. At this time point, RePAD immediately reports $v_7$ as an anomaly. The above detection process will repeat over and over again as time advances.

It is clear that the Look-Back parameter (i.e., $b$) is a critical parameter to RePAD since it controls the length of the preparation period and determines the detection performance of RePAD. In the next section, we investigate how this parameter impacts RePAD.

## 3  Evaluation Methodology

To comprehensively study the impact of the Look-Back parameter on RePAD, we consider the following six performance metrics:

1. Precision $(= \frac{TP}{TP+FP})$
2. Recall $(= \frac{TP}{TP+FN})$
3. F-score $(= 2 \times \frac{\text{Precision} \times \text{Recall}}{\text{Precision}+\text{Recall}})$
4. PP $(= 2b + 1$ time intervals$)$
5. LSTM retraining ratio $(= \frac{R}{N})$
6. Average detection time $\mu_d (= \frac{\sum_t^N T_t}{N-t+1})$ and standard deviation $\sigma_d (= \sqrt{\frac{\sum_t^N (T_t - \mu_d)^2}{N-t+1}})$

Recall that RePAD is capable of proactive anomaly detection, so using traditional well-known point-wise metrics (such as precision, recall, and F-score) to measure RePAD is inappropriate. Therefore, we propose an evaluation strategy based on [10] to capture both proactive detection performance and lasting detection performance. More specifically, if any anomaly occurring at time point $T$ can be detected within a detection time window ranging from time point $T - K$ to time point $T + K$, we say this anomaly is correctly detected, where $K \ll T$. We adopt the above-mentioned strategy together with precision, recall, and F-score to evaluate the detection performance of RePAD.

In the equations of precision and recall, TP, FP, and FN represent true positive, false positive, and false negative, respectively. Precision refers to positive predictive value, i.e., the proportion of positive/anomalous results that truly are positive/anomalous within the corresponding detection time windows. On the other hand, recall refers to the true positive rate or sensitivity, which is the ability to correctly identify positive/anomalous results within the corresponding detection time windows.

The F-score is a measure of a test's accuracy. It is defined as the weighted harmonic mean of the precision and recall of the test, i.e., F-score $= \left(\frac{2}{\text{Recall}^{-1}+\text{Precision}^{-1}}\right) = 2 \times \frac{\text{Precision} \times \text{Recall}}{\text{Precision}+\text{Recall}}$. The F-score reaches the best value at a value of 1, meaning perfect precision and recall. The worst F-score would be a value of 0, implying the lowest precision and the lowest recall.

PP, stands for Preparation Period, is a period of time required by RePAD to derive the required amount of AARE values for determining the detection threshold. More specifically, PP is defined as a time period starting when RePAD is launched and ending when RePAD officially starts its detection function. As mentioned earlier, RePAD starts at time point 0 and is able to derive the threshold at time point $2b + 1$, implying that PP would be $2b + 1$ time intervals. As $b$ increases, the corresponding PP will accordingly prolong, meaning that RePAD will need more time before it can start detecting anomalies in the target time series.

Recall that whenever time advances to the next time point, RePAD might need to retrain its LSTM model. If the retraining is required, such a process will prolong the time taken by RePAD to decide whether or not the data point collected at that moment is anomalous. In addition, RePAD might require more computation resources to retrain its LSTM model. Therefore, the LSTM-retraining ratio ($= \frac{R}{N}$) is another important indicator showing the impact of Look-Back parameter on RePAD, where $N$ is the total number of time points in the target time series, $R$ is the total number of time points at which RePAD has to retrain its LSTM, and $R \leq N$. If the ratio is low, it indicates several things. Firstly, it means that RePAD does not need frequent LSTM retraining since its LSTM model is able to accurately predict future data points. Secondly, it further means that RePAD is able to quickly make prediction and detection whenever time advances. Thirdly, it implies that RePAD does not need extra computational resources to retrain its LSTM. Therefore, the lower the ratio, the better efficiency RePAD is able to provide.

Since the goal of RePAD is to offer anomaly detection in real-time, it is essential to evaluate how soon RePAD is able to decide the anomalousness of every single data point in the target time series. Consequently, measuring average detection time $\mu_d$ required by RePAD and the corresponding standard deviation $\sigma_d$ are necessary. Hence, $\mu_d$ can be derived by equation $\frac{\sum_t^N T_t}{N-t+1}$, where $T_t$ is the time required by RePAD to detect if $v_t$ (i.e., the data point arrived at time point $t$) is anomalous or not, and $N$ as stated earlier is the total number of time points in the target time series. More specifically, $T_t$ is defined as a time period starting when time has advanced to $t$ and ending when RePAD has determined the anomalousness of $v_t$. Note that if RePAD needs to retrain its LSTM model at time point $t$, $T_t$ will also include the required LSTM retraining time. Apparently, if $\mu_d$ and $\sigma_d$ are both low, it means that RePAD is able to make detection in real-time.

In the next section, we investigate how the Look-Back parameter impacts RePAD based on all the above-mentioned metrics.

## 4     Evaluation Results

To investigate how the Look-Back parameter impacts RePAD, we designed four scenarios as listed in Table 1. In the first scenario, the Look-Back parameter is set to 3, following the setting in [1]. In the rest three scenarios, we separately increase the value to 30, 60, and 90. Note that we did not evaluate RePAD in scenarios where the value of the Look-Back parameter is less than 3 since RePAD under the settings will not have sufficient historical data points to learn the pattern of the target time series, leading to an extremely high LSTM retraining rate according to our experience. In addition, the network structure of LSTMs in the all scenarios are kept as simple as possible, i.e., one hidden layer with 10 hidden units.

Two empirical experiments were conducted based on two real-world time series datasets from the Numenta Anomaly Benchmark [11]. These datasets are called ec2-cpu-utilization-825cc2 and rds-cpu-utilization-cc0c53. They are abbreviated as CPU-cc2 and CPU-c53 in this paper, respectively. Table 2 lists the details of these two datasets. Note that the interval time between data points in all these datasets is 5 minutes. For this reason, RePAD in all the scenarios also followed the same time interval to perform anomaly detection. The two experiments were separately performed on a laptop running MacOS 10.15.1 with 2.6 GHz 6-Core Intel Core i7 and 16GB DDR4 SDRAM. Note that we followed [12] and set $K$ to 7 so as to determine the length of the detection time window. In other words, the detection time window is from time point $T - 7$ to time point $T + 7$ where $T$ is recalled as the time point at which an anomaly occurs. The performance metrics presented in Section 3 were all used in the experiments.

**Table 1.** RePAD in four scenarios. Note that $b$ is the look-back parameter.

| Scenario | 1 | 2 | 3 | 4 |
|---|---|---|---|---|
| The value of $b$ | 3 | 30 | 60 | 90 |
| Number of hidden layers | 1 | 1 | 1 | 1 |
| Number of hidden units | 10 | 10 | 10 | 10 |

**Table 2.** Two real-world time-series datasets used in the experiments.

| Dataset | Time Period | # of data points |
|---|---|---|
| CPU-cc2 | From 2014-04-10, 00:04 to 2014-04-24, 00:09 | 4032 |
| CPU-c53 | From 2014-02-14, 14:30 to 2014-02-28, 14:30 | 4032 |

### 4.1 Experiment 1

Figs. 2 and 3 show the detection results of RePAD on the CPU-cc2 dataset and a close-up result, respectively. Note that this dataset contains two anomalies labeled by human experts, and these two anomalies are marked as red hollow circles in both figures. It is clear that RePAD in Scenario 1 (i.e., $b = 3$) is the only one that is able to detect the first anomaly on time and the second anomaly proactively. In the rest of the scenarios, RePAD could not detect the first anomaly at all. In addition, RePAD in Scenario 1 has more true positives and less false negatives than RePAD in the other scenarios. This phenomenon can be clearly observed from Fig. 3, and this is the reason why the recall of RePAD in Scenario 1 reached the best value of 1 (as listed in Table 3). If we further take precision and F-score into consideration (see Table 3 as well), we can see that RePAD in Scenario 1 outperforms RePAD in the other scenarios, implying that setting the Look-Back parameter to be 3 seems appropriate for RePAD.

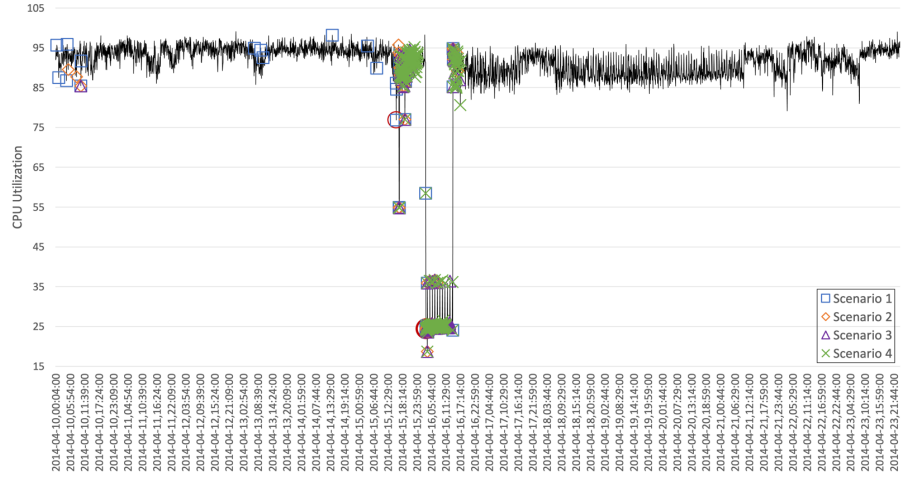

**Fig. 2.** The detection results of RePAD on the CPU-cc2 dataset. Note that this dataset has two labeled anomalies, marked as red hollow circles.

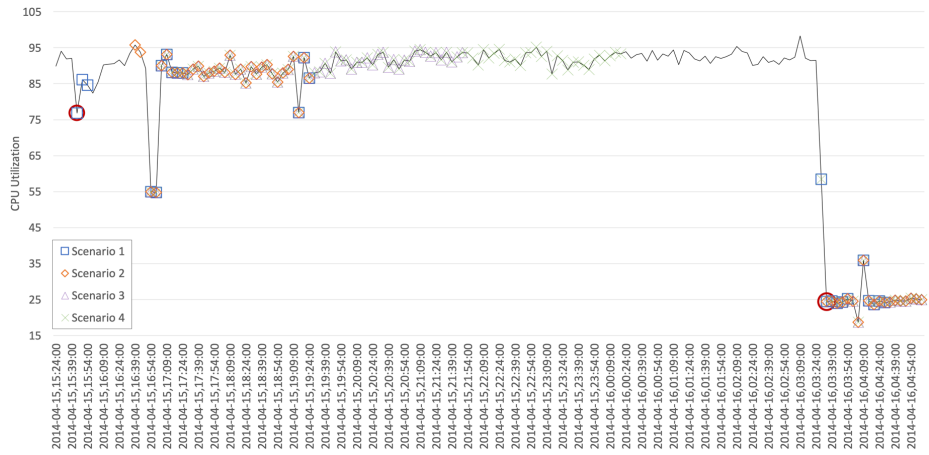

**Fig. 3.** A close-up of the detection results on the CPU-cc2 dataset. The two anomalies were marked as red hollow circles.

**Table 3.** The detection accuracy of RePAD on the CPU-cc2 dataset.

| Scenario | Precision | Recall | F-score |
|---|---|---|---|
| 1 | 0.4259 | 1 | 0.5973 |
| 2 | 0.0919 | 0.5 | 0.1553 |
| 3 | 0.0523 | 0.5 | 0.0947 |
| 4 | 0.0667 | 0.5 | 0.1176 |

Table 4 lists the PP required by RePAD on the CPU-cc2 dataset. In scenario 1 (i.e., $b = 3$), the required PP consists of only 5 ($= 2 * 3 - 1$) time intervals according to PP $= 2b + 1$. It means that after 7 time intervals, RePAD can officially start detecting anomalies in the time series. When we increased the value of the Look-Back parameter, the corresponding PP proportionally prolonged, implying that RePAD needed more time to get itself ready. In other words, RePAD loses its capability for providing readiness anomaly detection.

Table 5 summaries the retraining performance of RePAD on the CPU-cc2 dataset. We can see that RePAD in a scenario 1 only needs to retrain its LSTM model 83 times. The corresponding LSTM retraining ratio is very low since it is around 2% (=83/4027). However, when we increased the value of the Look-Back parameter, the number of LSTM retraining required by RePAD increased. Apparently, increasing the value of the Look-Back parameter leads to more frequent LSTM retraining. In other words, training the LSTM of RePAD with long-term historical data points does not help RePAD in learning the data pattern of the CPU-cc2 dataset and making accurate predictions. In addition, as Table 6 shows, the average detection time required by RePAD to decide the anomalousness of each data point increases when the value of the Look-Back parameter increases. The main reason is that the detection time includes both the corresponding LSTM retraining time (if the retraining is required) and the corresponding detection time. Since RePAD in Scenario 1 requires the least number of LSTM retraining and utilizes the least amount of training data (i.e., 3 data points), it leads to the shortest average detection time.

Based on the above results, we conclude that RePAD in Scenario 1 outperforms RePAD in the other three scenarios since it provides the best detection accuracy and time efficiency.

**Table 4.** The PP required by RePAD in different scenarios.

| Scenario | 1 | 2 | 3 | 4 |
|---|---|---|---|---|
| PP (Preparation Period) | 7 | 61 | 121 | 181 |

**Table 5.** The LSTM retraining performance of RePAD on the CPU-cc2 dataset.

| Scenario | # of required LSTM Retraining | LSTM Retraining Ratio |
|---|---|---|
| 1 | 83 | 2% |
| 2 | 119 | 2.97% |
| 3 | 199 | 5.01% |
| 4 | 257 | 6.5% |

**Table 6.** The time consumption of RePAD on the CPU-cc2 dataset.

| Scenario | Average Detection Time (sec) | Standard Deviation (sec) |
|---|---|---|
| 1 | 0.015 | 0.027 |
| 2 | 0.188 | 0.339 |
| 3 | 0.385 | 0.691 |
| 4 | 0.494 | 0.846 |

## 4.2 Experiment 2

In the second experiment, we employ RePAD in the four scenarios to detect anomalies on the CPU-c53 dataset. The detection results and a close-up result are shown in Figs. 4 and 5, respectively. Note that this dataset contains two anomalies labeled by human experts and marked as red hollow circles. Table 7 summaries the detection accuracy of RePAD. Apparently, RePAD in Scenario 1 is the only one that is able to detect the two anomalies within the detection time windows without generating many false positives, consequently leading to higher precision, recall, and F-score than RePAD in the rest of the scenarios. Note that both Scenarios 3 and 4 led to zero precision and recall, so it is impossible to calculate the corresponding F-scores.

When it comes to $PP$, RePAD on the CPU-c53 dataset requires exactly the same $PP$ as those as listed in Table 4. This is because the value of the Look-Back parameter is the only factor determining $PP$ (i.e., $2b+1$ time intervals). It is worth noting that the LSTM retraining ratio in this experiment does not always increase as the value of the Look-Back parameter increases. As listed in Table 8, RePAD in Scenario 3 requires the least number of LSTM retraining, whereas RePAD in Scenario 2 requires the highest number of LSTM retraining. In other words, the Look-Back parameter is not the only factor determining LSTM retraining ratio. The pattern of the target dataset is also influential. Nevertheless, we can see from Table 9 that the average detection time of RePAD increases when the value of the Look-Back parameter increases since the average detection time depends on not only LSTM retraining ratio, but also the amount of data points used to retrain the LSTM.

Overall speaking, Scenario 1 leads to the shortest $PP$ (Please see Table 4), the second least amount of LSTM retraining (see Table 8), and the shortest average detection time with the least standard deviation (see Table 9). In other words, configuring the Look-Back parameter to be 3 enables RePAD to promptly start anomaly detection and detect anomalies in real-time.

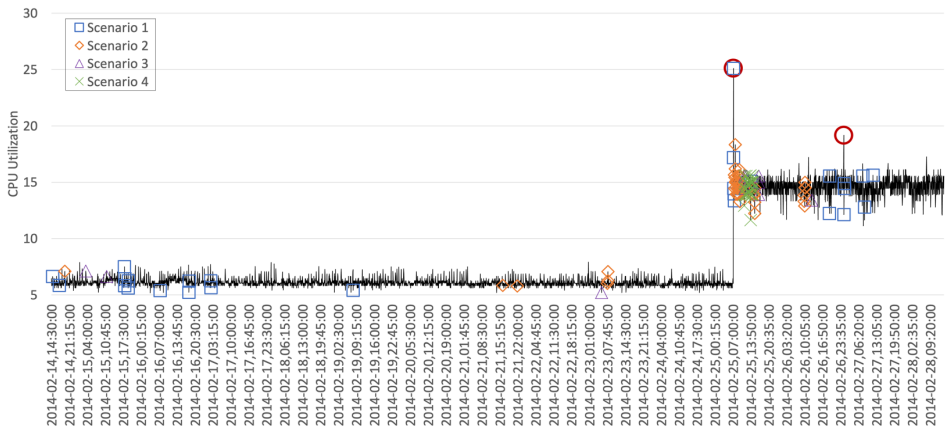

**Fig. 4.** The detection results of RePAD on the CPU-c53 dataset. Note that the two anomalies are marked as red hollow circles.

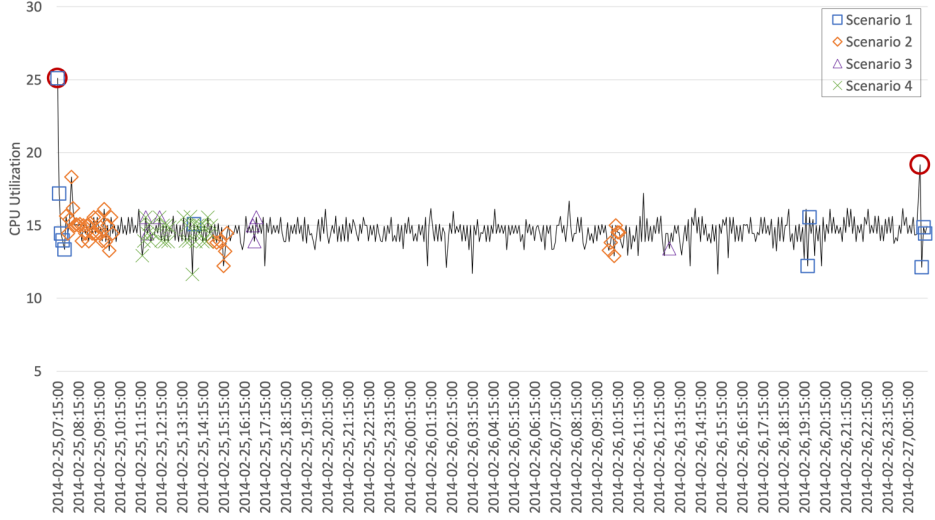

**Fig. 5.** A close-up of the detection results on the CPU-c53 dataset.

**Table 7.** The detection accuracy of RePAD on the CPU-c53 dataset.

| Scenario | Precision | Recall | F-score |
|---|---|---|---|
| 1 | 0.457 | 1 | 0.627 |
| 2 | 0.1569 | 0.5 | 0.2388 |
| 3 | 0 | 0 | n/a |
| 4 | 0 | 0 | n/a |

**Table 8.** The LSTM retraining performance of RePAD on the CPU-c53 dataset.

| Scenario | # of required LSTM Retraining | LSTM Retraining Ratio |
|---|---|---|
| 1 | 59 | 1.46% |
| 2 | 82 | 2.05% |
| 3 | 52 | 1.31% |
| 4 | 73 | 1.85% |

**Table 9.** The time consumption of RePAD on the CPU-c53 dataset.

| Scenario | Average Detection Time (sec) | Standard Deviation (sec) |
|---|---|---|
| 1 | 0.015 | 0.024 |
| 2 | 0.122 | 0.199 |
| 3 | 0.213 | 0.304 |
| 4 | 0.321 | 0.522 |

## 5 Conclusion and Future Work

In this paper, we have investigated how the Look-Back parameter influences the performance of RePAD by introducing a set of performance metrics that not only evaluate detection accuracy for novel proactive anomaly detection approaches, but also evaluate readiness and time efficiency. Regardless of which real-world dataset we utilized in our experiments, setting a lower value for the Look-Back parameter always leads to better performance on all the six considered metrics. More specifically, taking short-term historic data points as online training data is an appropriate strategy for RePAD since RePAD is able to promptly learn and adapt to the changing patterns of the target time series. Furthermore, using short-term historic data points also enables RePAD to soon start detecting anomalies in the target time series and detect potential anomalies in real-time. In all the scenarios discussed in this paper, scenario 1 (i.e., setting the Look-Back parameter to 3) is the most recommended since it enables RePAD to provide real-time, lightweight, and proactive anomaly detection for time series without requiring human intervention or domain knowledge.

As future work, we plan to improve RePAD by further reducing false positives. Furthermore, we would like to extend RePAD to detect anomalies on time series data observed from the eX$^3$ HPC cluster [13] by referring to [14][15][16] and designing the methodology in a parallel and distributed way.

**Acknowledgments.** This work was supported by the project eX$^3$ - *Experimental Infrastructure for Exploration of Exascale Computing* funded by the Research Council of Norway under contract 270053 and the scholarship under project number 80430060 supported by Norwegian University of Science and Technology.